\documentclass[runningheads]{llncs}
\usepackage{graphicx}
\usepackage{booktabs}
\usepackage[whole]{bxcjkjatype}
\usepackage{xcolor}
\usepackage[normalem]{ulem}
\usepackage[ruled,vlined]{algorithm2e}
\usepackage{subfigure}
\usepackage{amssymb, mathrsfs,amsmath}
\usepackage{cite}
\DeclareMathOperator*{\argmax}{arg\,max}

\graphicspath{ {figs/} }

\begin{document}
%
% \title{Adversarial Beats: Spoofed Arrhythmia in Automated ECG Diagnosis}
\title{Application of Adversarial Examples to Physical ECG Signals}

\titlerunning{Application of Adversarial Examples to Physical ECG Signals}
\author{Taiga Ono\inst{1} \and
Takeshi Sugawara\inst{2} \and
Jun Sakuma\inst{3} \and
Tatsuya Mori\inst{1,4}}
% \author{Taiga Ono \and
% Takeshi Sugawara \and
% Jun Sakuma \and
% Tatsuya Mori}
%
\authorrunning{T. Ono et al.}
\institute{Waseda University \and
The University of Electro-Communications \and
University of Tsukuba \and
RIKEN AIP \\}
% \institute{Waseda University, 3-4-1 Okubo Shinjuku 169-8555, Tokyo, Japan \and
% The University of Electro-Communications, 1-5-1 Chofu Chofugaoka 182-8585, Tokyo, Japan \and
% University of Tsukuba, 1-1ｰ1 Tennodai Tsukuba 305-8577, Ibaraki, Japan \and
% RIKEN AIP, 1−4−1 Chuo City Nihonbashi 103-0027, Tokyo, Japan \\}

\maketitle              % typeset the header of the contribution
\begin{abstract}
%（「The abstract should briefly summarize the contents of the paper in
% 15--250 words.ということなので、短縮が必要そうです。

%%% XXX 背景的な話はばっさりコメントアウトします 
% Recent advancements in clinical services powered by machine learning have been exposed to the threat of adversarial examples.
% While many researchers have developed machine-learning-based approaches that aim to diagnose patients based on various medical data, other researchers have demonstrated that adversarial examples can manipulate classification results. 
% \ono{Previous work on adversarial examples applied to medical classification tasks have demonstrated the effectiveness of numerically injecting adversarial perturbation directly into medical data, such as in the form of signals. How an attacker injects such adversarial inputs in a physical end-to-end setup, however, has remained an open research issue. The contribution of this work is to demonstrate the threat of a hardware attack that utilizes adversarial examples of electrocardiograms (ECGs) to induce misclassifications in cardiac diagnosis systems powered by machine learning algorithms. \sout{Furthermore, our research aims to shed light on the reality of adversarial examples of electrocardiograms (ECGs)}.}
%\onofix{Through our research, we aim to assess the reality and feasibility of adversarial examples for ECGs.}
% \mori{reality に shed light はちょっと婉曲的な気がするので、reality/feasibility を評価したと直接表現するのが良さそう}

This work aims to assess the reality and feasibility of the adversarial attack against cardiac diagnosis system powered by machine learning algorithms.
To this end, we introduce ``{\em adversarial beats}'', which are adversarial perturbations tailored specifically against electrocardiograms (ECGs) beat-by-beat classification system.
We first formulate an algorithm to generate adversarial examples for the ECG classification neural network model, and study its attack success rate.
Next, to evaluate its feasibility in a physical environment, we mount a hardware attack by designing a malicious signal generator which injects adversarial beats into ECG sensor readings.
To the best of our knowledge, our work is the first in evaluating the proficiency of adversarial examples for ECGs in a physical setup. 
Our real-world experiments demonstrate that adversarial beats successfully manipulated the diagnosis results 3–5 times out of 40 attempts throughout the course of 2 minutes.
% \sout{allowing the hardware attack to} successfully feign the existence of abnormal cardiac beats. 
Finally, we discuss the overall feasibility and impact of the attack, by clearly defining motives and constraints of expected attackers along with our experimental results.

% \sout{we consider the feasibility and impact of adversarial attacks against ECG classifiers in the real world.}}
% \mori{この最後のセンテンス意図がちょっとわかりにくいです。feasibility があると結論づけたのか否か}

\keywords{Deep Learning  \and Adversarial Examples \and Hardware Security.}
\end{abstract}
\section{Introduction}
\label{sec:intro}

The application of neural networks in clinical diagnosis processes has gained popularity in recent years. For instance, there have been numerous studies applying pattern recognition on digital medical images such as X-rays and CT scans of patients to spot irregularities such as indicators of strokes or tumors~\cite{dnn-imaging-applications}.
Conventionally, such a diagnosis requires trained physicians to spend an extended period of time analyzing measurements, where as neural networks are able to quickly and accurately spot suspicious patterns. 
Machine-learning-powered services specializing in clinical applications~\cite{Shah2019, company:enlitic, company:arterys} have been on the rise lately, and the United States Food \& Drug Administration has given neural networks clearance for use in medical services~\cite{fda-clearance}. Neural networks are not limited to image classification tasks as Pranav~et~al.~\cite{Pranav2017} have demonstrated that neural networks can be leveraged to discover signs of arrhythmia in electrocardiograms (hereinafter referred to as ECGs).

While neural networks show great promise in automated clinical diagnosis, it could be exposed to the threats of {\em adversarial examples}~\cite{Szegedy2014}. 
% \mori{\sout{XXX add more AE references here XXX}}
Past work shows that a trained classifier to diagnose medical images can be fooled by medical images perturbed by adversarial noise, causing classifiers to make mistakes in diagnosing illnesses~\cite{Finlayson2019}. Recent studies~\cite{Kurakin2016, Goodfellow2015, Athalye2017, Chen2019, Yakura2019, Tu2020} have also explored the robustness of the adversarial examples in the physical world.
There are a few existing studies on generating adversarial perturbations applied on ECGs~\cite{Han2019, Chen2019}, where generated perturbations are injected into ECG measurements to cause misclassifications in segment-based discovery of atrial fibrillation.
% \sout{These studies leveraged convolution neural network tailored for image classification tasks to study the effectiveness of the adversarial examples against ECGs. XXX ←これ正しい?XXX}
% \ono{上記は先行研究の論文に書いてあった内容とは異なっていたので線を引いております}
% To the best of our knowledge, however, no prior studies have assessed the feasibility of such adversarial attacks against cardiac diagnosis systems from the viewpoints of real-world constraints and plausible attack scenarios. 
We note, however, that these previous studies have been limited to software simulations, with limited discussion on the feasibility of adversarial examples of ECG in the real world. In addition, difficultly in realizing adversarial examples via noise injection into sensors have been reported as well, due to noise effects~\cite{Yakura2019, Carlini2018}. 

Given this background, this work aims to bridge the gap between the threats assessed by the software simulation and those that may arise in the real world. We introduce ``{\em adversarial beats}'', which are adversarial perturbations tailored specifically against ECG, taking into account the physical constraints in implementing the attack.
%We assess the feasibility of {\em adversarial beats} in the real world, both in terms of the physical constraints in implementing the attack, and the plausible scenarios that motivate attackers to perform the attack.}
Our work explores the following research questions: \\
\textbf{RQ1:} In the real world, what types of attackers would be motivated to leverage {\em adversarial beats} against an ECG diagnosis system, under what constraints? \\
\textbf{RQ2:} Can we generate {\em adversarial beats} against machine-learning-powered ECG diagnosis systems to alter classification results to lead to a meaningful manipulation of ECG diagnosis results? \\
\textbf{RQ3:} Can we apply the generated {\em adversarial beats} for ECGs in a hardware attack, taking into account physical constraints?
% \sugawara{Maybe RQ3 should come first so that we can claim that we designed the algorithm/hardware for those specific constraints.}
%To pursue these research questions, we propose a PoC attack against an automated ECG classification scheme. Understanding the means for an adversary to leverage adversarial examples in hardware manipulation requires detailed analysis of the specific classification task that the target neural network is assigned to, along with a deep understanding of the specifics of the particular field the task pertains to. 
% We believe that exploring the incentives and limitations posed specifically by the clinical field \ono{and its unique data-acquisition process and various actors,} offers insight into the nature of how an adversary formulates an attack utilizing adversarial examples against clinical classification tasks. 

Key contributions of our work are summarized as follows: We first provide a plausible implementation scheme of neural networks in automated clinical diagnostics of medical data acquired from patients via monitoring, which without proper precautions could be vulnerable to manipulation. We also clarify the types of attackers that may be motivated to attack such implementations, along with their methods and constraints. We then introduce {\em adversarial beats}, which aims to spoof classification results of ECG diagnosis. A success rate of up to 65.4\% is achieved during training. Finally, we perform real-world evaluation of the {\em adversarial beats} through a PoC hardware attack. 3-5 successful attack cases are achieved throughout 40 attempts. Our analysis gives in-depth insight into the feasibility of the attack as well as the realistic scenarios in which adversaries find clear incentives to perform the attack with {\em adversarial beats}. We hope future system designers refer to our work to review the threat of real-world attacks leveraging adversarial examples on ECG diagnosis systems.
% \begin{mybullet2}
%      \item We provide a plausible implementation scheme of neural networks in automated clinical diagnostics of medical data acquired from patients via monitoring, which without proper precautions could be vulnerable to manipulation.
%     \item We introduce {\em adversarial beats}, which aims to spoof classification results of ECG diagnosis.
%     A success rate of up to 65.4\% is achieved during training.
%     \ono{\item We clarify the types of attackers 
%     % \sout{and defenders} (文字数が余っていたら防御者について書きたいと思います) 
%     that may be motivated to use or detect {\em adversarial beats}, along with their methods and constraints.}
%     \item We perform real-world evaluation of the {\em adversarial beats} through a PoC hardware attack. 3-5 successful attack cases are achieved throughout 40 attempts.
%     \item We give in-depth insight into the feasibility of the attack as well as the realistic scenarios in which adversaries find clear incentives to perform the attack with {\em adversarial beats}. 
% \end{mybullet2}
\section{Background: ECG Monitoring and Classification}
% and Related Work}
\label{sec:background}

In this section, we present the basics of ECG monitoring and describe the ECG heartbeat classification model we adopt in this work.

%\mori{next 以降がなくなるので、first をとりましょうか}
% \remove{Next, we present our threat model considering the following: the target system, the attack process, motivation of adversaries, and the capability of the adversary.}

\subsection{ECG monitoring}

% \mori{メモ: この節では、ECG}

ECGs are waveforms commonly used to visualize and monitor the electrical activity of a patient's heart over time, allowing physicians to spot certain patterns entailing potential health risks.
One of such patterns are arrhythmias, which is a broad class of irregular heartbeats~\cite{ecg-workout}. While most arrhythmias are considered to be harmless, some indicate signs of dangerous heart activities which could lead to fatal conditions~\cite{ecg-workout}.
While Holter monitors~\cite{Galli2016} have been a conventional means to monitor ECGs over a extended period of time, recent development of wearable medical devices~\cite{Turakhia2013} offers convenient, non-intrusive methods of monitoring patient ECGs, allowing physicians to analyze heart activities of patients throughout their daily lives.
As more convenient means of ECG monitoring develop, however, measured patient data will increase, along with the demand for an efficient and accurate analysis of measured ECGs.

\subsection{ECG Classification task}

% % 心電図の波形説明画像
% \begin{figure}
% \centering
% \includegraphics[width=0.3\textwidth]{figs/qrs.pdf}
% \caption{Component Peaks of a Heartbeat ECG.} \label{fig:qrs}
% \end{figure}

%In this section, we present a target model of an ECG diagnosis system
% In this section, we describe the ECG heartbeat classification model we adopt in this work.
% \remove{, adhering to the details of the target system shown in Section~\ref{sec:background:threat}}
% . We first describe in detail the data pre-processing schemes. 
% We then 
%\ono{Next, we present our threat model considering the following: the target system, the attack process, motivation of adversaries, and the capability of the adversary. We take into consideration the actors within a clinical environment, and how they may be motivated to attack such a system, under what sort of constraints.}

% \subsection{\remove{Data pre-processing} \ono{Target ECG Classification Model}}
\label{sec:target_ecg_class_model}

\paragraph{\bf Heartbeat Classification Task}

% Table for ANSI/AAMI EC57 Standards
\begin{table}
\centering
\caption{Five Heartbeat Classes Specified in ANSI/AAMI EC57~\cite{AAMI}.} \label{tbl:aami}
\begin{tabular}{cc}
\toprule
Heartbeat Class & Descriptions\\ \midrule
N& Average/Normal\\
S& Atrial/Nodal premature\\
V&  Ventricular premature\\
F&  Fusion of V and N\\
Q&  Paced/Unclassifiable\\
\bottomrule
\end{tabular}
\end{table}

% \mori{\sout{最初に NN で心拍を分類するアプローチが増えてきているという話を書きましょう。その上で、下記のような分類、データが標準的に使われているという話をします}}

Convolutional neural networks
% offer a potential solution to the aforementioned increase in measured ECG data to diagnose. Methods suggested in previous work 
are capable of categorizing individual heartbeats extracted from an ECG~\cite{Kachuee2018}, or detecting signs of atrial fibrillation from an arbitrary segment of an ECG ~\cite{Pranav2017}.
To evaluate the performance of arrhythmia classification algorithms, it is essential to have a standard protocol/dataset for researchers/engineers to compare the results. 
Association for Advancement of Medical Instrumentation (AAMI) recommends specific protocols for evaluating performance of automated ECG classifiers, and ANSI/AAMI EC57~\cite{AAMI} specifies the five types of heartbeats in Table~\ref{tbl:aami}:
non-ectopic beats (N), supra-ventricular ectopic beats (S), ventricular ectopic beats (V), fusion beats (F) and unclassifiable beats (Q), where N is considered normal heartbeats, S, V, and F are considered arrhythmic beats.

\section{Threat Model}
\label{sec:threat}

% \remove{Target ECG Classification Model} 
% \ono{Threat Model \& Identifying Goals}
\label{sec:target_model}

In this section, we lay out the target system of our proposed attack. To this end, we consider how machine learning-based ECG classifiers would be generally implemented in clinical settings in the future. Furthermore, we identify potential adversaries in such clinical settings.

% \mori{この章内容が多いので、整理したほうが良さそうです。具体的には 3.1, 3.2はモデルの話なので、4章に移すなど}

% \subsubsection{\ono{Assumptions Made on the Target Model}}

% \ono{\sout{While other forms of attacks against neural networks are showcased in piror work (poisoning, etc., 関連論文記載する), this research focuses on the capabilities of adversarial examples. Thus, we assume attackers may have access to varying (depending on the attacker's profile) parameters of the neural network which they can leverage to craft their adversarial examples, but won't directly manipulate the neural network model itself.}}

\subsection{A Model of Target Clinical Diagnosis System}

% \remove{The infrastructure with which neural networks are implemented in automated clinical diagnosis systems remains an open research topic.}
% \ono{The infrastructure with which ECG classifiers explained in section~\ref{sec:target_ecg_class_model} are implemented in automated clinical diagnosis systems remains an open research topic.}

% As such, 

\begin{figure}[tbp]
\centering
\includegraphics[width=0.5\textwidth]{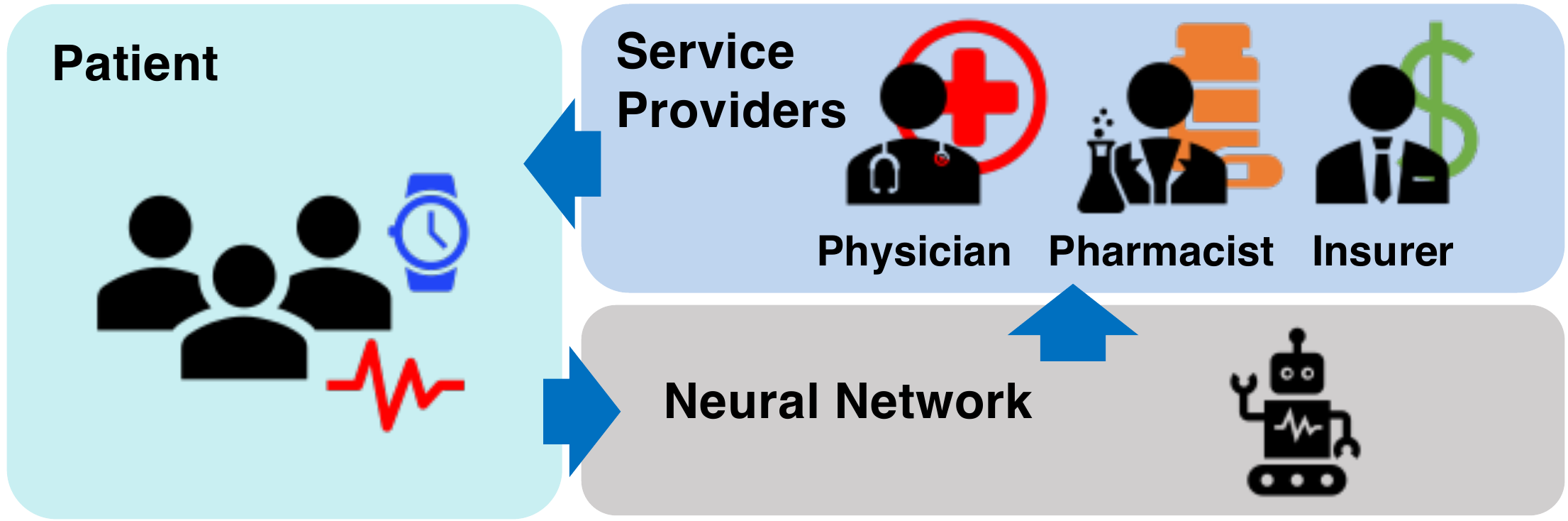}
\caption{A model of the Target Clinical Diagnosis System.} \label{fig:target-system}
\end{figure}

There is a significant incentive in building fully-automated clinical diagnosis system, as it offers a potential means to cutting down healthcare expenses~\cite{Finlayson2019}. 
% Thus, an ideal automated diagnosis system is expected to operate without a designated operator, requiring only the input data to output a diagnosis.
Analyzing how exactly a potential attacker can manipulate a neural network-based diagnosis system is not a straightforward process, because there is no single diagnosis system infrastructure to consider. 
We thus propose a reference model, considering what is to be expected from future clinical diagnosis systems.
% To illustrate the threat model, we suggest a target system, which an institution may build and deploy in the future. 
Our reference model shown in Figure~\ref{fig:target-system} revolves around daily monitoring of patient ECGs, aimed towards catching signs of suspicious heart activity entailing potential health risks. 
These tasks are expected to become increasingly convenient if classifiers are implemented in conjunction with newer IoT medical wearables, allowing physicians to analyze large amounts of monitored data and spotting any irregularities quickly. ECGs measured from IoT wearables on patients are sent to a central classifier trained to diagnose the collected data, which the results are sent to healthcare providers to base their decisions on. 

%This model leads to an assumption that without preventive measures, malicious service providers and/or technicians may be able to insert adversarial contraptions inside measurement devices to manipulate measurement results. While it is unclear whether or not the classifier network is unique to medical institutions or standardized across a larger scale, we expect countermeasures to be implemented against direct data/model parameter tampering~\cite{Finlayson2019}. 

% 標的診断システムの説明画像

% \subsection{Heartbeat Spoofing Attack Models}
% \subsubsection{Attack Motivation}

%The National Health Care Anti-Fraud Association estimates that yearly tens of billions of dollars are lost by healthcare fraud in the United States~\cite{nhcaa}. Adversarial examples are applicable in medical fraud, allowing adversaries to manipulate input data to control the results of a diagnosis without direct tampering of the measured data or the classifier parameters, and influencing resulting clinical decisions~\cite{Finlayson2019}. An example of fraudulent clinical transactions is increasing the billed amount of expenses for a given patient by purposely performing unnecessary procedures, done by healthcare providers with monetary incentives~\cite{nhcaa}. 

\subsection{Potential Adversaries}

% \sugawara{*\sout{It seems that we implicitly assume false-negative attacks. We need to explain we focus on a particular one maybe after explaining the big picture.}*}

We now identify potential adversaries that could take advantage of the proposed system with malicious intent. We highlight their motives, as well as their abilities to go about an attack, allowing us to identify potential constraints for each adversary.
All but one of the suggested attackers have the potential of conducting our proposed attack. We note, however, that multiple types of attackers could conspire together and share profits earned, also potentially lifting constraints that otherwise would be imposed if attacking on their own. \\
{\bf Patient: }
Malicious patients could be incentivised to avoid expenses from procedures or medication, by faking diagnosis results to their favor, such as by feigning normal heartbeats to mask arrhythmias (false-negative). Although in the long run, it is the patient's health at risk, it is possible that those with monetary incentives attempt such an attack, leading to negative impacts on the patient themselves, and other actors involved in the process. We note that, while we conduct experiments to launch a false-positive attack of arrhythmia discovery, false-negative attacks would be possible by setting different target classes. 
% We leave experimenting false-negative attacks as future work. 
% As for their abilities, 
Patients will spend some arbitrary time with the target IoT device, if not for a prolonged period depending on its application. Though limited in the extent of tampering the device, they do have physical access to the target device itself. \\
{\bf Hospital Personnel: }
Hospital personnel could be physicians, nurses, or any other personnel in charge of patient care. 
% While one would assume that profit gained from care of a patient does not go directly to the healthcare workers, 
Malicious personnel could be incentivised to conspire against a patient, manipulating diagnosis results to incur additional fees for unnecessary procedures and checkups. 
% Though it may depend on the specific positions of personnel, 
Hospital personnel would have arbitrary access to maintain devices, including IoT devices, giving them just as much access to target devices as patients, if not more. 
% It should be noted, however, that physical access to target devices will be limited once the IoT device is installed on the patient.
 \\
{\bf Medical IoT Manufacturers: }
IoT manufacturers would be considered a part of the supply chain, in which devices that they manufacture are then utilized by hospitals and other client facilities. Although indirectly, such manufacturers may be motivated to conspire with hospitals for monetary incentives.
% While manufactured devices are expected to go through various checks and audits before seeing use in a real clinical environment, 
Manufacturers would have the most extensive access to the device itself, enabling advanced tampering on the device. Attacks via hardware trojans installed within the supply chain is a known attack vector explored in previous work~\cite{Kan2016}. \\
{\bf Third-Party Healthcare Organizations: }
Third-party healthcare organizations refer to organizations such as pharmacies and insurance companies. Their profits and services depend heavily on diagnosis results made by hospitals, and malicious actors within such organizations could be incentivised to conspire with other service providers for their own gain. It should be noted, however, that these organizations would have very limited physical access to the target device itself, making it very difficult for them to act on their own. This makes such malicious parties a co-conspirator, rather than the actual attacker, but with potentially the most monetary incentive out of all other attacker profiles.
% \sugawara{The several potential attackers are presented, but it's not clear which one we are going to consider. At least we should mention where we will discuss the correspondence between the threat model and the attack requirements.}

The proposed types of potential attackers will be discussed in section~\ref{sec:discussion}, when they will be considered whether or not it will be feasible for them to go about with the proposed real-world attack method.
% \subsection{\ono{Attacker Knowledge}}
In addition to the abilities of the attacker, we consider what they know about the target system. While it is unclear how neural network parameters are disclosed to the public in a future clinical diagnosis system, we assume that critical parameters are undisclosed to the public, given recent advancements in machine-learning security~\cite{Papernot2018}. Studies regarding ECG classification tasks, however, often utilize datasets available online, which adversaries could also obtain easily. This leads us to believe that all types of adversaries, including patients, are capable of conducting a black-box attack on the classifier. While it is unclear how clinical institutions will regulate how ECGs are diagnosed by neural networks, we assume that healthcare providers are capable of white-box attacks on such a system, provided that they have insider access to such information, such as the exact means of which ECG is recoreded, preprocessed, and fed into classifiers.
% \sugawara{*There are several recent side-channel attacks for stealing the ML parameters. Maybe we can cite them for justifying the secret-parameter model.*}

\section{Adversarial Beats}
\label{sec:adversarial_beats}

% ノイズの載せられた波形の画像
\begin{figure}
\centering
\includegraphics[width=0.6\textwidth]{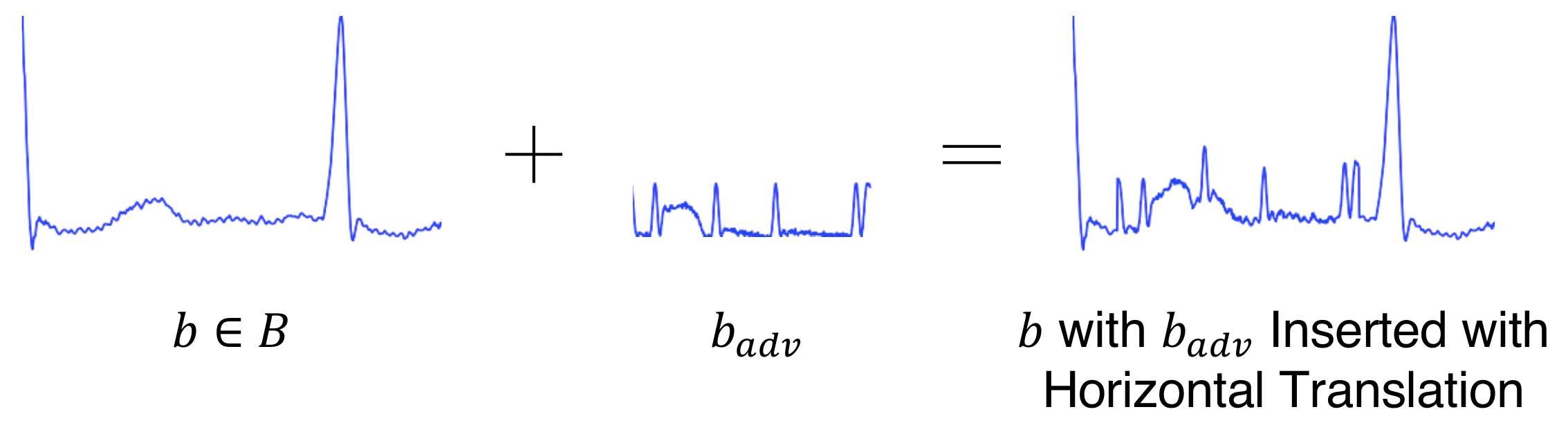}
\caption{Inserting {\em Adversarial beats} in Target ECG. 
% \sugawara{Please explicitly mention that this is the concrete example (if it is)}
}
\label{fig:ab-on-ecg}
\end{figure}

This section covers the principles and methods we implement to generate {\em adversarial beats} against an ECG diagnosis system. 
We present the physical constraints for achieving the real-world attack, and describe the {\em adversarial beats} generation algorithm against the underlying heartbeat classifier.

\subsection{Overcoming Physical Constraints}
\label{sec:adversarial_beats_constraints}

{\em Adversarial beats} must overcome physical constraints posed by the ECG pre-processing pipeline. 
To ensure that {\em adversarial beats} functions in the physical realm, the following challenges are to be considered: \\
% \sout{To make our attack model work in the real world, we need to tackle the following challenges:}
{\bf Real-Time Attack:}
Existing studies~\cite{Han2019, Chen2019} attempted to generate specific adversarial perturbations for each sample, similar to the studies of conventional adversarial examples. From a physical standpoint, however, such an approach is not feasible due to the fact that the signals are to be generated in real-time, as ECGs are measured from a patient.\\
{\bf Generating Short Noises:}
% \sout{Universal} 
Adversarial perturbations generated in previous work~\cite{Chen2019} are around 30 seconds long at its longest, to approximately 5 seconds long, with diminishing effectiveness the shorter it gets, for certain classes at the point of direct digital injection. 
Longer noise makes managing where to inject it in the ECG difficult, and may suffer from arbitrary changes in the target ECG (such as heart rate/patient movement) during the noise injection, which is assumed to have an influence on its effectiveness. This leads us to believe that shorter perturbations increase the chances of a successful attack via physical injection. Thus, {\em adversarial beats} are optimized to be shorter and robust to shifting within the range of a single heartbeat. 
% \ono{Due to this, however, amplitude limitation is the most taxing on the effectiveness of the {\em adversarial beats}, given a small window to generate noise. We consider the increase in amplitude from previous work to be a necessary cost of realizing real-world attacks.} 
\\
{\bf Existence of Physical Band-pass Filters:}
To make the adversarial perturbations work in the real-world, we need to consider the existence of band-pass filters implemented in the preprocessing stage before classification, which filters out any frequency components in the measured signal outside of a certain range. Conventionally, raw ECG segments are processed through certain bandpass filters to remove unwanted noise artifacts after measurement, commonly caused by improper electrode placements, external devices, or patient movement~\cite{Kher2019}. The specifications of noise artifacts considered throughout this work are summarized in Table~\ref{tbl:freq-spec}. Although Chen~et~al.~\cite{Chen2019} addressed the issue by applying a frequency limit on the generated perturbations, they did not evaluate its effectiveness in a physical environment. Conventional ECGs are sampled at around 300--360 Hz to ensure no information is lost~\cite{Kher2019}. For {\em adversarial beats} to maintain their effectiveness after filtering, they must be constrained to the range of frequencies that the filter allows. As done in the work of Chen~\textit{et~al.}\cite{Chen2019}, we limit the frequency components of the {\em adversarial beats} within the range target ECGs are filtered with. {\em Adversarial beats} are recorded to the extent of its sampling rate, and thus should have granularity no greater than 300--360 Hz. \\

% ノイズの周波数テーブル
\begin{table}[tbp]
\caption{Effects and Frequencies of Common Noise Artifacts in ECGs~\cite{Kher2019}.}\label{tbl:freq-spec}
\centering
{\footnotesize
\resizebox{\columnwidth}{!}{%
\begin{tabular}{cccc}
\toprule
Noise Type & Frequency~[Hz] & Causes & Effects\\
\midrule
Baseline Wandering & 0.5 & Electrode Placement & Vertical Displacement\\
Powerline Interference & 50--60 & External Devices & Sinusoidal interference\\
Motion Noise & 1--10 & Patient movement & Artifacts mistaken as QRS complex\\
\bottomrule
\end{tabular}
}
}
\end{table}

{\noindent \bf Existence of Heartbeat Segmentation:}
% ECG前処理の段階説明画像
% \begin{figure}
% \centering
% \includegraphics[width=0.3\textwidth]{figs/ecg-pipeline.pdf}
% \caption{Sequential Pipeline for ECG Processing.} \label{fig:ecg-pipeline}
% \end{figure}
% Figure~\ref{fig:ecg-pipeline} presents a procedural diagram showing the preparation pipeline of measured ECGs for diagnosis.
Another component in the ECG pre-processing pipeline is the heartbeat segmentation operation. While there are various methods to diagnose an ECG, we implement beat-by-beat classification--a classification scheme to diagnose individual heartbeats in segments of ECGs. We believe it best represents diagnosis of ECGs that are monitored throughout daily life, as signs of arrhythmia throughout daily activities are indicators of health risks. Due to this fact, we propose that considering beat-by-beat classification a distinction from previous work, as it introduces a different impact compared to segmentation-based classification, which classify entire segments of ECGs to a limited range of classification types. Without any regard for its waveform, {\em adversarial beats} can substantially alter the general waveform of the ECG regardless of the implemented filter, which is assumed to be the result of strict constraints. Injected intrusive {\em adversarial beats} drastically alter the waveform of the ECG from its original state, causing the beat segmentation algorithm to detect additional nonexistent heartbeats. Discrepancies in numbers of detected heartbeats can be the source of arbitrary errors outputted by the classifier. Thus, in addition to detection from human perception, {\em adversarial beats} must avoid detection from such heartbeat segmentation algorithms. Preliminary experiments lead to the conclusion that limiting the amplitude of the {\em adversarial beats} is the simplest method of preventing interference with segmentation. \\
% A waveform for a single heartbeat consists of the P and T waves, and a central QRS complex, comprised of 3 components: the Q-wave, the R-peak, and the $\mathcal{S}$-wave (not to be confused with the S used to describe a heartbeat class)~\cite{ecg-workout}, as shown in Figure~\ref{fig:qrs}. R-peaks are the most prominent peak of the QRS complex~\cite{ecg-workout}, making them a suitable point of reference for individual heartbeats. To segment measured ECGs into individual heartbeats, our target model implements QRS detection algorithms to map out the expected locations of R-peaks in a measured ECG. 
% While there are multiple known methods for QRS detection~\cite{Hamilton2002, Engelse1979, Christov2004}, 
% our work abstracts various QRS detection methods down to general constraints applied to {\em adversarial beats}, obtained from heuristic analysis.\footnote{Due to page constraints, readers are directed to the following sources~\cite{Hamilton2002, Engelse1979, Christov2004} for example implementations of heartbeat segmentation algoirthms.}
% As we show in the next section, preliminary experiments lead to the conclusion that limiting the amplitude of the {\em adversarial beats} is the simplest method of preventing interference with QRS detection. We note that while this experiment only considered the existence of beat segmentation algorithms in the preprocessing methods, other preprocessing methods may be implemented to ensure quality of measurements. We leave the potential effects of additional processing methods as points to consider for future work.
{\bf Universal perturbations:}
% \sugawara{The term universal perturbations appears in advance at "Generating Short Noises".}
Physical adversarial perturbations need to be {\em universal} in the following two aspects:
(1) Beat Invariance: They must be valid for any heartbeat.
There is uncertainty in what the class of the original measured heartbeat pertains to. To maximize the success of spoofing certain classes of heartbeat under any occasion, {\em adversarial beats} are trained to be effective on any types of heartbeats, and not just normal ones.
(2) Positional Invariance: They need to be valid anywhere in the ECG signal.
Similar to past work, the concept of Expectation Over Transformation~\cite{Athalye2017} is applied to {\em adversarial beats} to be effective regardless of where in the target ECG it is applied, as aiming the injection at an exact relative location in an ECG is difficult for an attacker. Specifically, this technique involves implementing random degrees of horizontal shift when applying {\em adversarial beats} to an ECG signal~\cite{Chen2019}. This is because it is infeasible for hardware to inject {\em adversarial beats} in an exact position on the ECG signal in a physical implementation. This contributes to the positional-invariance, or the universal characteristic regarding injection position. \\

\subsection{Adversarial Beats Generation Algorithm}

With the physical constraints shown above, we generate {\em adversarial beats} by training them on the preprocessed dataset until the misclassification proficiency ceases to increase. They are optimized solely on a digital environment, utilizing datasets available to the public, with digital transformations to simulate physical constraints. To optimize the amplitude of the {\em adversarial beats} without compromising its effectiveness, we implement an iterative algorithm shown in Algorithm~\ref{alg:gen}, located in the appendix.
To {\em adversarial beats}, we adopt the ``Expectation Over Transformation,'' proposed by Athalye~\textit{et~al.}~\cite{Athalye2017}.
The algorithm aims to generate adversarial examples that remain adversarial over a chosen transformation in the physical world.
It has been adopted in several studies such as Chen~\textit{et~al.}~\cite{Chen2019} and Brown~\textit{et~al.}~\cite{Brown2017} for generating robust adversarial examples.

The optimization problem for generating {\em adversarial beats} is formulated with Equation~\ref{eq:optimization}, which aims to minimize the categorical cross-entropy loss for the targeted class while minimizing the frequency components of the {\em adversarial beats} that will be filtered out during the physical ECG signal processing stage. In summary, this optimization equation optimizes {\em adversarial beats} to adapt to all target heartbeats and horizontal translation, while minimizing the amplitude of frequency components outside of the designated frequency range.
\begin{equation}
b_{adv}^{\ast}=
   \argmax_{b_{adv}} \{\mathbb{E}_{b\sim \mathcal{B}, x\sim \mathcal{X}}
    \left[\log P\left(y_t\mid b + T(b_{adv}; x)\right)\right] 
     - \lambda\sum_{f} {\left| M(f)\cdot\mathscr{F}(f; b_{adv})\right|}
    \}
\label{eq:optimization}
\end{equation}
% \sout{where,} 
$\mathcal{B}$ is the set of all heartbeat ECG segments in the training set, $\mathcal{X}$ is the set of all possible horizontal shift transformations applicable on the {\em adversarial beat} $b_{adv}$ within the heartbeat, and $y_t$ is a targeted heartbeat class. 
Note that $x\sim \mathcal{X}$ refers to a random variable following a uniform distribution on $\mathcal{X}$.
$T(\cdot)$ is an operator that represents the horizontal shift transformation, i.e., placing the {\em adversarial beat}, $b_{adv}$, within the target heartbeat ECG segment. A similar approach is taken in previous work~\cite{Chen2019}.
$\mathscr{F}(f; x(t))$ represents a frequency component of a given signal $x(t)$; it is computed by applying the Fourier transform to $x(t)$. $M(\cdot)$ represents a function to represent the effect of the bandpath filters; i.e., it applies a mask such that frequency components that are to be filtered out during the ECG processing pipeline remain; i.e., $M(f) = 1$ if the frequency $f$ is to be filtered out, otherwise  $M(f) =0$. During optimization, a capacity on the amplitude of the {\em adversarial beat} waveform is set to be contained within 0--$A$, where $A$ is specified by Algorithm~\ref{alg:gen} in the appendix.

The resulting {\em adversarial beat} $b_{adv}^{\ast}$ causes an ECG segment to be missclassified as a target class when inserted. 
Figure~\ref{fig:ab-on-ecg} illustrates an example case of an {\em adversarial beat} being inserted into a target heartbeat ECG segment. Since the set of target heartbeat ECG segments used to optimize the {\em adversarial beats} contain heartbeats of any class, they are applicable to heartbeats of any class. This contributes to the universal characteristic of {\em adversarial beat} regarding heartbeat classes. We note that the {\em adversarial beat} is shorter in length compared to the target ECG heartbeat segment.
The main difference from the optimization model used by Chen~\textit{et~al.} is that the regularization by $L_2$-norm disregarded, as we are focused on avoiding interference with the segmentation algorithm. 
Furthermore, the {\em adversarial beats} are intended to be inserted on every heartbeat that the adversary intends to spoof classification results for.
{\em Adversarial beats} tend to have greater amplitude and less smoothness compared to those  reported in previous work~\cite{Chen2019, Han2019}, which have been demonstrated only in the digital environment. Similar tendency can be observed in physical adversarial perturbations in other domains, such as image~\cite{Eykholt2018, Brown2017} and audio~\cite{Yakura2019}.
The resulting {\em adversarial beats} from this optimization algorithm have optimized amplitudes to minimize heartbeat segmentation interference, and are robust to ECG frequency filtering.
\section{Simulation-based Evaluation}
\label{sec:result}

% \sugawara{Maybe we should change the section title to "Simulation-based Evaluation" to emphasize how it's different from the next section.}

% % AB-Sをinjectしたヒートマップ画像
% \begin{figure}
% \includegraphics[width=0.5\textwidth]{figs/S-AP-misclas-heatmap-new.png}
% \caption{Cases of Misclassifications by Digital Injection (AB-S).} \label{fig:digital-injection-heatmap-abs}
% \end{figure}

% % AB-Vをinjectしたヒートマップ画像
% \begin{figure}
% \includegraphics[width=0.5\textwidth]{figs/V-AP-misclas-heatmap-new.png}
% \caption{Cases of Misclassifications by Digital Injection (AB-V).} \label{fig:digital-injection-heatmap-abv}
% \end{figure}

In this section, we begin by evaluating the proficiency of the trained target heartbeat classifier.
% in detecting various classes of heartbeats. 
We then generate {\em adversarial beats} using our proposed algorithm, and evaluate its effectiveness in spoofing certain classes of heartbeats by digitally inserting them in ECG segments.

\subsection{Target Heartbeat Classifier}
\label{sec:result:target}

At its core, the ECG diagnosis system is expected to take in a segment of an ECG of a patient and automatically return diagnosis results on them. Several algorithms are applied on the ECG sequentially as discussed in section~\ref{sec:adversarial_beats_constraints}, outputting a series of data that the target neural network classifier can properly analyze. In our implementation, we filter out relevant frequencies for baseline wandering and powerline interference using forward-backward filtering with cut-off frequencies allowing frequencies of 0.5--50~Hz. Frequencies to attenuate artifacts caused by patient movement are considered and implemented during heartbeat segmentation. Because the target neural network analyzes the waveform of ECGs of individual heartbeats, the measured ECG data must be segmented into individual heartbeats. Several heartbeat segmentation algorithms have been proposed~\cite{Hamilton2002, Christov2004, Engelse1979}. The detected heartbeats are sliced from the original ECG segment and adjusted to fit the specified input of the classifier with their values normalized with 0--1. We implement heartbeat segmentation by adopting algorithms proposed by Hamilton~\cite{Hamilton2002}.
% \sugawara{There are basic (and important) info about filtering and classification, which should come earlier, possibly in Section 4.1.}
% \mori{上記菅原先生のコメントにフォローして。これらは、過去文献では考慮されていなかった physical constraints であると説明すると良いでしょうか。これらを考慮することが新規性である、と位置づけられると思います。heartbeat segmentation については、threat model に沿った攻撃を実現するには segmentation-based ではだめで、beat-by-beat が必須である、というような説明で良いと思います。スペースの都合で、sequential pipeline の図は割愛でも良さそうです。}
% \ono{上記コメントを受けまして4.1に基礎的な説明を移しました。実際すでに4.1には似たような説明がありましたので、ここにあった文章を4.1と混ぜ、必要に応じて追記いたしました。なお、こちらに残った文は4.1の知識を踏まえて実験に使った分類機をどう実装したかの説明になっております。}
% CNNの構築説明画像
% \begin{figure}[tbp]
% \centering
% \includegraphics[width=0.3\textwidth]{figs/network-structure.pdf}
% \caption{Classifier Architecture.} \label{fig:model}
% \end{figure}
% % 攻撃対象の分類機のCM画像
\begin{figure}
\centering
\includegraphics[width=0.3\textwidth]{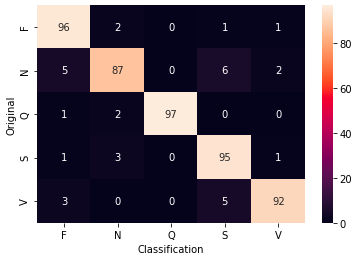}
\caption{Confusion Matrix of Target Classifier.} \label{fig:ecg-heatmap}
\end{figure}
As the baseline neural network model, we adopt the ECG heartbeat classification model developed by Kachuee~et~al.\cite{Kachuee2018}. Details of the architecture and training datasets are specified in the appendix.

The confusion matrix for the target classifier on the test dataset consisting of 100 heartbeats for each class sampled from the test dataset is shown in Figure~\ref{fig:ecg-heatmap}. 
% \sugawara{\sout{We need to briefly explain how to read the confusion matrix.}}
Each axis represents classes of heartbeat, and the X-axis represents the classification result, while the Y-axis represents the ground truth. Each tile represents the number of resulting classifications.
The overall accuracy of the trained classifier was 93.4\%. The classifier is mostly able to consistently distinguish between different classes of heartbeats. Overall, the classifier made consistent classifications on ECGs measured from a test subject, suggesting translation of the classification proficiency on data measured during the experiments.

\subsection{Generating Adversarial Beats}

Two variations of {\em adversarial beats} each serving unique purposes are trained. Training data and evaluation data prepared in the previous section is used to
generate the {\em adversarial beats}, similar to the target classifier. The first {\em adversarial beat} is trained to cause the target classifier to misclassify any heartbeat injected with it as S class heartbeats, regardless of their original class and spoofing an S class heartbeat (hereinafter referred to as AB-S). The second {\em adversarial beat} is trained similarly so that any injected heartbeats are misclassified by the target classifier as V class heartbeats, regardless of their original class, spoofing a V class heartbeat (hereinafter referred to as AB-V). Specifications of the two {\em adversarial beats} are shown in Table~\ref{tbl:used-aps}. AB-S achieved acceptable accuracy of up to 65.4\% chance of successfully causing a targeted misclassification,  with a relatively low amplitude of 0.1875. On the contrary, AB-V required a higher amplitude of 0.4 and adjustments to its length to achieve acceptable accuracy of 56.2\% chance of causing a targeted misclassification. These results were achieved through heuristic adjustments of the $\rho$ parameter from Algorithm~\ref{alg:gen}. While a relatively acceptable degree of effectiveness and amplitude was achieved by threshold value $\rho = 0.65$ for AB-S, there was difficulty in generating AB-V with similar effectiveness without having to maintain large amplitude. $\rho$ for AB-V was then decreased, which resulted in acceptable amplitude. This suggests S class heartbeats are easier to spoof compared to V class heartbeats, thus requiring less perturbations to cause a misclassification. 

% 生成されたABの情報テーブル
\begin{table}[tbp]
\caption{Specifications of Generated {\em Adversarial Beats}.}\label{tbl:used-aps}
\centering
{\setlength{\tabcolsep}{2em}
\begin{tabular}{cccc}
\toprule
Target Class & $\ell_{\infty}$ & $\ell_{2}$ & Success Rate [\%]\\
\midrule
S (AB-S) & 0.1875 & 1.44 & 65.4\\
V (AB-V) & 0.4000 & 5.51 & 56.2\\
\bottomrule
\end{tabular}
}
\end{table}

% ヒートマップの画像集
\begin{figure}[t]
    \centering
    % \subfigure[No Injection]{\includegraphics[width=.32\linewidth]{figs/model-9-prediction-heatmap.png}\label{fig:heatmaps-none}}
    \subfigure[Injected AB-S]{\includegraphics[width=.3\linewidth]{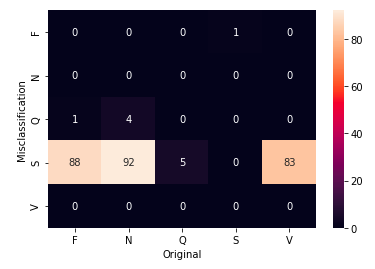}\label{fig:heatmaps-abs}}
    \subfigure[Injected AB-V]{\includegraphics[width=.3\linewidth]{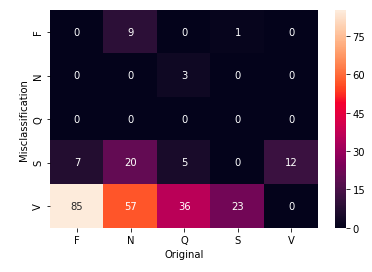}\label{fig:heatmaps-abv}}
    \caption{Confusion Matrix of Target Classifier Under Different Injection
    % \sugawara{We should improve the image quality}
    }
    \label{fig:heatmaps}
\end{figure}

\subsection{Testing Adversarial Beats}

The trained {\em adversarial beats} are tested by introducing them into the test set and counting the cases of misclassification by the target classifier. This is done in a digital environment, isolated from the hardware experimentation to follow. Our intent is to first ensure that the generated {\em adversarial beats} possess a certain degree of effectiveness at this point. To simulate the uncertainty in where the {\em adversarial beat} may be injected during hardware injection and test whether they generalize to various translation, {\em adversarial beats} are digitally injected with random shifts throughout the sampling axis in the original heartbeat ECG.
Figure~\ref{fig:heatmaps-abs} and Figure~\ref{fig:heatmaps-abv} presents cases of misclassifications. Each axis represents classes of heartbeat, and the X-axis represents the ground truth, while the Y-axis represents the classification result. Unlike Figure~\ref{fig:ecg-heatmap}, tiles in Figure~\ref{fig:heatmaps-abs} and Figure~\ref{fig:heatmaps-abv} represent cases of misclassifications. AB-S and AB-V both show acceptable performance, as they cause expected instances of misclassifications according to the accuracy denoted throughout training.
\section{Real-World Experiments}
\label{sec:hardware}

%\mori{ハードウェア構成を示す図、写真を復活させましょう}

% \sugawara{\sout{This section is too short. Possibly integrate them with the section header?}}
In this section, we apply the generated {\em adversarial beats} into a physical attack, and perform a physical hardware-based attack against our representation of a ECG diagnosis system. We perform a wired-signal injection to the device used for ECG measurement. Hardware setup is explained, and proficiency of the attack is reported with the introduced metrics.

\subsection{Experimental Setup and Procedure}

% ハードウェアセットアップの説明画像
\begin{figure}[tbp]
    \centering
    \subfigure[Concept Diagram]{\includegraphics[width=.50\linewidth]{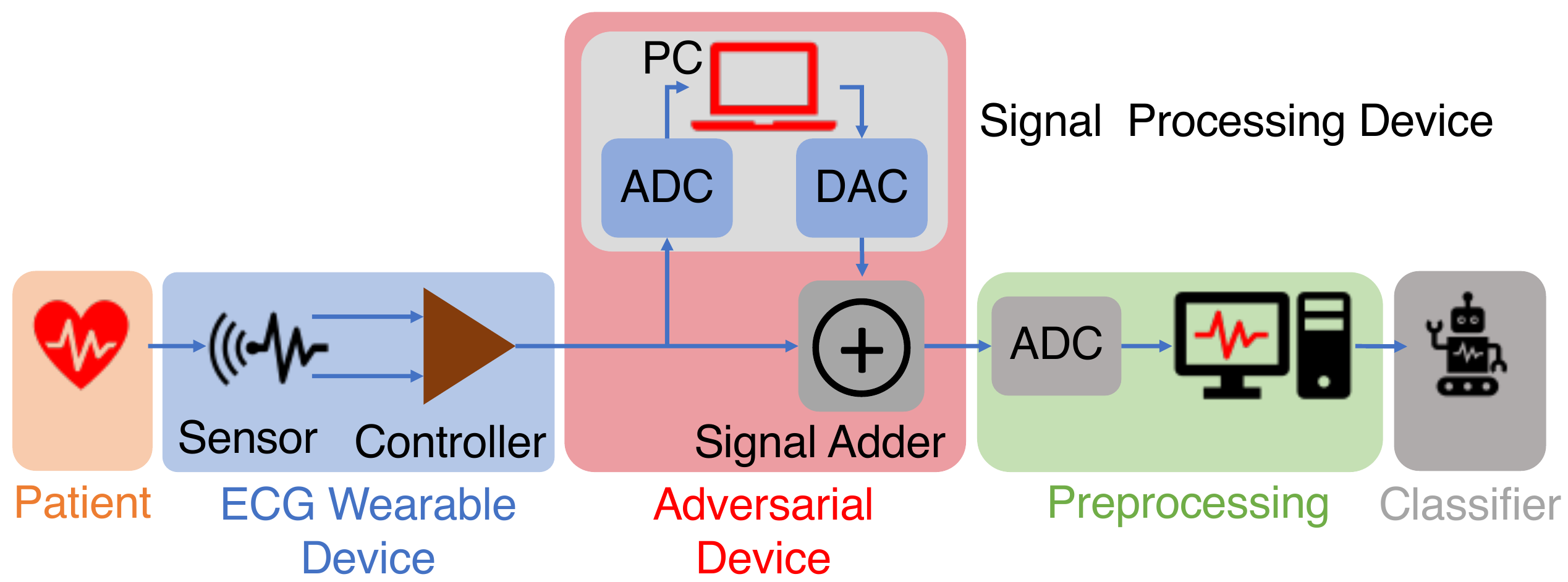}\label{fig:hardware-setup}}
    \subfigure[Photo]{\includegraphics[width=.35\linewidth]{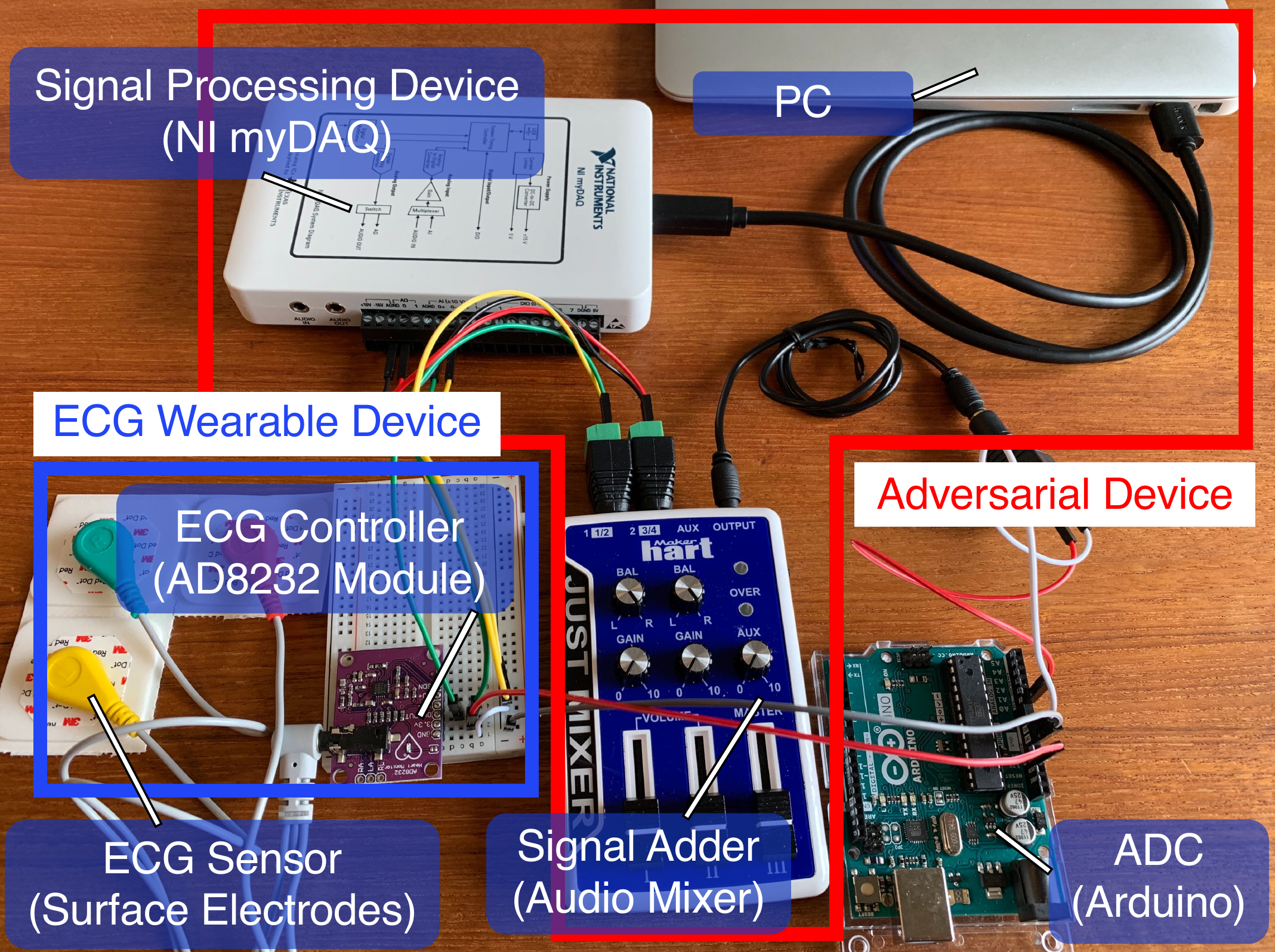}\label{fig:hardware-setup-photo}}
    \caption{Poc Hardware Setup.}
    \label{fig:hardware-setups}
\end{figure}

% % inject前のECG波形画像
% \begin{figure}
% \includegraphics[width=0.5\textwidth]{figs/non-injected-ecg-fix.png}
% \caption{Sample Segments of Experimental ECGs (No Injection).} \label{fig:injected-seg-before}
% \end{figure}

% % inject前のECG波形画像
% \begin{figure}
% \includegraphics[width=0.5\textwidth]{figs/s-injected-ecg-fix.png}
% \caption{Sample Segments of Experimental ECGs (Injected with AB-S).} \label{fig:injected-seg-after}
% \end{figure}

The diagram of our hardware setup is shown in Figure~\ref{fig:hardware-setup}. 
% The setup contains components needed for constructing a target ECG measurement device. 
In our setup, a conventional surface electrode is attached to the chest of a research participant, to emulate an arbitrary ECG measurement device. An ECG controller module reads the measured signals and performs amplification and filtering to the raw signal, outputting an unperturbed ECG as an analog signal.
This raw ECG signal is then fed into a signal processing device, consisting of an analog-to-digital converter (ADC), a digital-to-analog converter (DAC), and a PC\footnote{We discuss the feasibility of hardware implementation in Section~\ref{sec:discussion}.}. The signal processing device reads the unperturbed ECG signal for a given period to monitor the timing to inject an {\em adversarial beat}. The signal processing device has an {\em adversarial beat} pre-computed from the methods covered in section~\ref{sec:adversarial_beats}~and~\ref{sec:result}, and transmits a segment of given length, containing an arbitrary number of the prepared {\em adversarial beats} separated by the computed interval. The injection waveform generated by the signal processing device (considered as the malicious device in our attack) and the unperturbed ECG signals are then combined via signal addition. Finally, the resulting combined signals are digitized at the target device's ADC, and sent to our ECG classifier. The resulting PoC hardware setup is shown in Figure~\ref{fig:hardware-setup-photo}. It's specifics are explained in the appendix. We note also, that the hardware used for the setup is easily obtainable in terms of accessibility and expenses.

\noindent{\bf Signal Processing}

The signal processing device executes an algorithm to output {\em adversarial beats} in pulses that synchronize with the patient's heartbeat, ensuring accurate injection. The {\em adversarial beat} is also scaled so that it is injected in the intended amplitude relative to the ECG it is injected in. This algorithm ensures that {\em adversarial beats} are aligned to heartbeats, so that they retain their attack success rate regardless of which heartbeat it is injected in. The algorithm executed is as follows:
\begin{enumerate}
    \item[1)] Read the ECG from the target patient for 5 seconds.
    \item[2)] Filter the ECG and perform heartbeat detection.
    \item[3)] Compute the distance between individual heartbeats that are detected.
    \item[4)] Compute the amplitude of the measured ECG.
    \item[5)] Scale the amplitude/length of the pre-computed {\em adversarial beat} to match the amplitude of the measured ECG/length of measured heartbeat distance.
    \item[6)] Construct a 5 second signal, comprised of duplicates of the scaled {\em adversarial beats} padded by the calculated heartbeat distances so that they occur in the same rhythm as the ECG.
\end{enumerate}

\noindent{\bf Procedure}

The proposed hardware-injection attack is executed while measuring the ECG of a human participant, who has their heart activity measured for 120 seconds each trial. Before the experiment, we obtained informed consent and confirmed that the participant had no heart defects or disorders, i.e., all the ECG segments measured should be classified as N. {\em Adversarial beats} are injected for a total of 40 times every trial, and the injected ECG is filtered before undergoing beat segmentation. Here, we use the trained classifier from section ~\ref{sec:result:target}. Finally, the segmented ECGs with individual heartbeats are fed into the classifier by batch, outputting the predicted class of every heartbeat in the measured ECG. The instances of each detected beat classes are recorded to show how often the proposed hardware attack was able to spoof the targeted heartbeat class. In our experiments, the following ECGs were measured, with a total of 10 trials performed for each: ECGs without any injections as control, ECGs injected with AB-S, and ECGs injected with AB-V.

\subsection{Injection Attack Result}

% Inject後の診断結果テーブル
\begin{table}[tbp]
\centering
\caption{Diagnosis Results for ECGs.}\label{tbl:injection-proficiency}
{\setlength{\tabcolsep}{1.5em}
\begin{tabular}{crrrrr}
\toprule
Target Class & F & N & Q & S & V\\
\midrule
None (Control) & 0 & 138.6 & 0 & 11.6 & 0\\
S & 0 & 124.2 & 0 & 16.4 & 0\\
V & 0.1 & 112.6 & 0.4 & 44.9 & 3.1\\
\bottomrule
\end{tabular}
}
\end{table}

\begin{figure}[tbp]
    \centering
    \subfigure[No Injection]{\includegraphics[width=.35\linewidth]{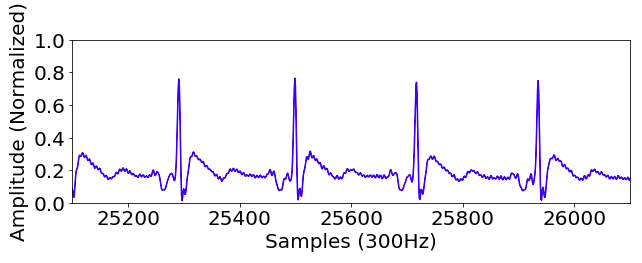}\label{fig:injected-seg-before}}
    \subfigure[Injected with AB-S]{\includegraphics[width=.35\linewidth]{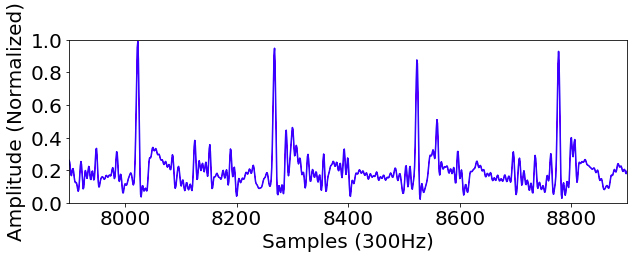}\label{fig:injected-seg-after}}
    \caption{Sample Segments of Experimental ECGs.}
    \label{fig:injected-seg-before-after}
\end{figure}

% \mori{最終的に攻撃成功確率がどうなるかの話をもう少し詳しく書く。3-5 success / 40 attempts はどこかあら？}

Table~\ref{tbl:injection-proficiency} summarizes the measurement results averaged over the 10 separate trials. A normal ECG waveform, and an ECG waveform measured with AB-S injected during measurement is displayed in Figure~\ref{fig:injected-seg-before-after}.
When counting the additional arrhythmic beats discovered from injected ECGs compared to ECGs with no injection, we see that we were able to spoof 3-5 cases of heartbeats (approxmiately 5 additional S-class arrhythmia for ECGs injected with AB-S, and 3 additional V-class arrhythmia for ECGs injected with AB-V). This suggests that the {\em adversarial beats} are capable of spoofing certain heartbeats, generalize to unseen samples, and retain their effectiveness in a physical environment. We found that in the control sample, several heartbeats were misclassified as S class, resulting in a classification error rate of roughly 7.8\%. While the original performance of the target classifier was suspected to be the prime cause of additional S class heartbeat detection, S class heartbeats are considered to possess a unique attribute, as the classifier also seldom misclassified average N class heartbeats as F class beats, which did not occur during the hardware-based attack. Additional observations are included in the appendix due to page limit.

% Appendixへ移動
% An unexpected result during our hardware experiment was how AB-V was more proficient at spoofing S class heartbeats than AB-S. AB-V requires a higher amplitude and longer length overall to reach a lower effectiveness threshold, suggesting the V class heartbeats to be more difficult in spoofing. As seen in the digital evaluation of the {\em adversarial beats}, AB-V tended to cause N class heartbeats to be seldom misclassified as S class heartbeats. Thus, AB-V is assumed to also be capable of spoofing S class heartbeats as an unintended side-effect. Small artifacts in the waveform that are essential to its effectiveness may have been lost through hardware noise and preprocessing, lowering its ability to spoof V class heartbeats. This leaves only its ability to spoof S class heartbeats, which is assumed to require less detail, considering a much smaller amplitude of the {\em adversarial beat} is required to achieve threshold frequency for spoofing S class heartbeats. Besides the aforementioned side-effect, we see that the two {\em adversarial beats} were able to spoof heartbeats of the intended classes up to 3-5 times within 40 tries of injections.
\section{Discussion}
\label{sec:discussion}

\subsection{Feasibility of the Attack}

We discuss the feasibility of the attack, considering the cost, integration, and installation of the hardware required for performing the attack with the {\em adversarial beats}.
We also take into consideration our suggested attacker models, and consider the feasibility of our proposed real-world attack in their perspective.

\noindent{\bf Attacker's Knowledge}

As mentioned in section~\ref{sec:threat},
% all types of 
attackers are expected to be capable of conducting a black-box attack on the target ECG system. Our experimental results based on a white-box attack shows that the success rate is not necessarily very high,
% \sout{(XXX 具体的な数字を追記)}
as we are only capable of spoofing 3-5 additional arrhythmic heartbeats in a physical setting. As white-box attacks generally have a higher success rate compared to black-box attacks, we conclude that a successful black-box attack in a physical setting has a limited threat model, leading us to believe that malicious hospital personnel are best fit to utilize {\em adversarial beats}, 
% \scout{(その理由を追記).}
as they are the only actors who are presumed to have remotely any access to confidential information to the target ECG classification model.

\noindent{\bf Cost}

Through our experiments we see that creating an {\em adversarial beat} is cheap in terms of resources, requiring little resource commitment from the adversary. As for the hardware used in the injection attacks, the adversary only needs to prepare the injection components of the hardware used in this experiment, i.e., a computer, ADC/DAC, and an audio mixer, which all are available at reasonable expenses. 
This is significant from the perspective of malicious patients or malicious hospital personnel, where advanced hardware tampering consisting of expensive hardware is infeasible. Our PoC hardware setup is possible for any adversary with reasonable knowledge and resourcefulness. This is, however, not to say that an advanced adaptation of our PoC is pointless.

\noindent{\bf Integration}

With the growing trend in medical IoT and their expected convenience, we expect daily biomedical monitoring devices to be small in size. We believe that with the simplicity of our injection algorithm, the PoC hardware setup we propose can be embedded into a much smaller circuit component by a resourceful adversary, making potential tampering with the monitoring hardware unnoticeable. We note that the components used in our experimental setup are prototypes of the proposed attack, and integrating all these components into a small board, using a microcontroller with ADC/DAC and op-amps, is possible for an adversary with moderate expertise in embedded systems.
Malicious medical IoT manufacturers would be capable of such tampering. With additional functionality, such as generating adversarial beats when prompted via a network signal, {\em adversarial beats} can be leveraged to poison a supply chain, enabling a type of backdoor that malicious manufacturers and their co-conspirators can utilize to manipulate diagnosis results.

\noindent{\bf Installation}

% \remove{The hardware injection attack implemented in our experiments should be possible for malicious healthcare providers, as they are expected to be the ones providing the measurement devices for the patient to use. Service providers may be able to install the adversarial hardware contraption in the manufacturing phase.}
The installation of malicious hardware could occur in different phases depending on the attacker. As malicious hospital personnel are expected to be the ones maintaining measurement devices before providing them to the patient, they could install the adversarial contraption anytime before provision. On the contrary, patients will have arbitrary access to the measurement device during the extent of the measurement being taken. Malicious manufacturers may install the adversarial hardware contraption in the manufacturing phase. For malicious patients and hospital personnel, it is straightforward to attach an adversarial hardware circuit to an ECG monitoring device as we demonstrate in this work. If such tampering is undetected by other stake holders, the device can be used as an adversarial ECG measuring device, which outputs perturbed heartbeats by command, but otherwise returns the original, unperturbed ECG of the patient. 

\noindent{\bf Overall Feasibility}

Taking the aforementioned details into consideration, it can be concluded that our proposed attack requires hospital personnel with access to the physical target device and insider knowledge of the target ECG neural network classifier to succeed.
While the exact degree of knowledge hospital personnel would have on the classification model is unclear, our proposed attack could be possible in the case of lack of access restriction to model parameters. We hope that future system designers can refer to our research as an example of a possible threat to look out for.

\subsection{Limitations}

We discuss further points of optimization and context which could improve our proposed hardware attack.

\noindent{\bf Adversarial Beats}

There is much room for optimization of the {\em adversarial beats}. Their lengths are manually adjusted as a static value at the beginning of training, resulting with a heuristically optimized parameter at best. Improvement to the {\em adversarial beat} generation algorithm to find the optimal length of the {\em adversarial beat} is prioritized in future work. We note, also, that the waveform of {\em adversarial beats} are optimized to minimize the disruptiveness of the generated {\em adversarial beats} towards beat segmentation algorithms. While experimental results have shown heuristically that minimizing the prominence of the peaks in {\em adversarial beats} sufficiently prevents beat segmentation disruption, the specific conditions in which beat segmentation algorithm mistake {\em adversarial beats} as QRS complexes are unclear. Thus, specific beat segmentation algorithms should be considered in detail for future work to analyze the specific constraints they put on {\em adversarial beats}.
% Appendixに移動
% Besides optimizing the waveform to minimize disruptiveness, results of our experiments warrant improvement in the optimization function for training {\em adversarial beats}. In this work, {\em adversarial beats} were only trained to increase the chances of spoofing a target class and not particularly trained to avoid spoofing heartbeats to any other classes. This in turn introduced a side effect of {\em adversarial beats} spoofing unintended heartbeat classes when they failed to spoof the true target heartbeat class. Additional constraints should be introduced to prevent {\em adversarial beats} from converging towards spoofing unintended heartbeat classes.
An increase in the overall effectiveness of {\em adversarial beats} would also lead to a larger variety of potential attackers to consider. Improving on the generation algorithm of {\em adversarial beats} to enable a black-box attack will assist in expanding the discussion of a more diverse threat model.

\noindent{\bf Hardware Implementation}

While the hardware setup showcased in this work demonstrates the feasibility of physical adversarial examples for ECGs, implementation could be improved. Besides hardware noise, the injection algorithm implemented in the myDAQ device may also be responsible for the inaccuracy of recreating {\em adversarial beats} as the injection signal. As the algorithm only accommodates for difference in lengths of particular heartbeats by scaling the length of {\em adversarial beats} by average beat-to-beat distance within a 5-second window, the resulting length adjustment may be inaccurate, which can be taxing on their effectiveness. Furthermore, while {\em adversarial beats} are trained to be effective in any horizontal position on a target heartbeat, ending up on a prominent peak in the waveform significantly alters the maximum amplitude of the resulting injected ECG, which introduces a degree of uncertainty that could be detrimental to the success of the attack. Improvements in generating injection signals may allow adversaries for more consistent and versatile manipulation of the diagnosis results.

% 割愛
% This work also only elaborated on the feasibility of direct wired signal injection, which limits the threat model to direct tampering of the measurement device via hardware manipulation. Different mediums of signal injections such as wireless EMI injections~\cite{Kune2013, Chen2019} can potentially enable a broad demographic of adversaries to execute attacks in various methods with different incentives. Our work offers a prototype of one of such various attacks, to ensure the possibility of more sophisticated attacks that expand the possibilities of the attackers. Future work will include considerations of different hardware apparatus to execute the hardware injection attack.

\noindent{\bf Target System}

Our proposed attack is naive in the sense that it disregards the presence of human operators in the automated diagnosis system. {\em Adversarial beats} introduce prominent noise into the measured ECG signals. While we assume an automated diagnosis, a specialist in ECGs simply observing the measured ECG is enough to raise suspicion and potentially compromise the hardware attack. In this case, our human observer would be the hospital personnel, making them even more of a likely candidate to carry out the proposed attack.
Future work focuses on minimizing the amplitude of {\em adversarial beats}, or otherwise overcoming the risk of the attack getting compromised by a human observer.

\subsection{Ethical Consideration}
The hardware implementation used in this work was not used for any purposes other than for experimentation. We obtained informed consent from the participant who had their ECGs measured; i.e., before the experiments, we explained the objectives of the experiments and that we use the measured ECGs purely for the experiments, and not for other purposes such as medical diagnosis.

\section{Conclusion}
\label{sec:conclusion}

Automated clinical diagnosis systems powered by neural networks are exposed to the threats of adversarial examples. To analyze the feasibility of adversarial examples for ECGs in the real world, we attempt a PoC attack against a naively implemented neural network-based automatic ECG diagnosis system. Using off-the-shelf hardware to implement our attack, we demonstrate that our proposed hardware attack is capable of causing
% surface electrodes placed on patients' chests to read their heart activity as adversarial ECGs, causing 
a classifier trained to detect arrhythmia to make a misclassification. We also identify attackers and scenarios in which the proposed attack could be executed to commit a healthcare fraud. As this work showed, the specific domain knowledge and the specific schemes in which measurement data is collected play a substantial part in how an adversary may leverage adversarial examples in the real world. 
We hope our work can be referenced by system designers to implement preventive measures against potential adversarial attacks in future clinical diagnostic systems. 

% \sugawara{The reference list is still broken.}

\newpage
\bibliographystyle{splncs04}
\bibliography{ref}

\begin{thebibliography}{10}
\providecommand{\url}[1]{\texttt{#1}}
\providecommand{\urlprefix}{URL }
\providecommand{\doi}[1]{https://doi.org/#1}

\bibitem{fda-clearance}
Administration, U.F..D.: Artificial intelligence and machine learning in
  software as a medical device.
  \url{https://www.fda.gov/medical-devices/software-medical-device-samd/artificial-intelligence-and-machine-}
  \url{learning-software-medical-device}

\bibitem{Athalye2017}
Athalye, A., Engstrom, L., Ilyas, A., Kwok, K.: Synthesizing robust adversarial
  examples. In: Proceedings of the 35th International Conference on Machine
  Learning. Proceedings of Machine Learning Research, vol.~80, pp. 284--293
  (July 2018)

\bibitem{Brown2017}
Brown, T.B., Man{\'{e}}, D., Roy, A., Abadi, M., Gilmer, J.: Adversarial patch.
  CoRR  \textbf{abs/1712.09665} (2017)

\bibitem{Carlini2018}
Carlini, N., Wagner, D.: Audio adversarial examples: Targeted attacks on
  speech-to-text. In: Proceedings of the IEEE Symposium on Security and Privacy
  (2018)

\bibitem{Chen2019}
Chen, H., Huang, C., Huang, Q., Zhang, Q.: {ECGadv: Generating Adversarial
  Electrocardiogram to Misguide Arrhythmia Classification System}. CoRR
  \textbf{abs/1901.03808} (2019), \url{http://arxiv.org/abs/1901.03808}

\bibitem{Christov2004}
Christov, I.: Real time elecrocardiogram {QRS} detection using combined
  adaptive threshold. Biomedical engineering online  \textbf{3}, ~1--9
  (September 2004). \doi{10.1186/1475-925X-3-28}

\bibitem{Engelse1979}
Engelse~WAH, Z.C.: A single scan algorithm for {QRS}-detection and feature
  extraction. Computers in Cardiology  \textbf{6},  37--42 (1979)

\bibitem{Eykholt2018}
Eykholt, K., Evtimov, I., Fernandes, E., Li, B., Rahmati, A., Tram{\`{e}}r, F.,
  Prakash, A., Kohno, T., Song, D.: Physical adversarial examples for object
  detectors. In: 12th {USENIX} Workshop on Offensive Technologies, {WOOT} 2018.
  pp. 1--10 (2018)

\bibitem{Finlayson2019}
Finlayson, S.G., Kohane, I.S., Beam, A.L.: Adversarial attacks against medical
  deep learning systems. CoRR  \textbf{abs/1804.05296} (2018),
  \url{http://arxiv.org/abs/1804.05296}

\bibitem{Galli2016}
Galli, A., Ambrosini, F., Lombardi, F.: Holter monitoring and loop recorders:
  From research to clinical practice. Arrhythmia \& Electrophysiology Review
  \textbf{5}(2),  136--143 (August 2016). \doi{10.15420/AER.2016.17.2}

\bibitem{Goldberger2000}
Goldberger, A., Amaral, L., Glass, L., Hausdorff, J., Ivanov, P., Mark, R.,
  Mietus, J., Moody, G., Peng, C.K., Stanley, H.: {PhysioBank},
  {PhysioToolkit}, and {PhysioNet} : Components of a new research resource for
  complex physiologic signals. Circulation  \textbf{101}(23),  215--220 (July
  2000). \doi{10.1161/01.CIR.101.23.e215}

\bibitem{Goodfellow2015}
Goodfellow, I., Shlens, J., Szegedy, C.: Explaining and harnessing adversarial
  examples. In: International Conference on Learning Representations (2015),
  \url{http://arxiv.org/abs/1412.6572}

\bibitem{Hamilton2002}
Hamilton, P.: Open source {ECG} analysis. Comput Cardiol  \textbf{29},
  101--104 (October 2002). \doi{10.1109/CIC.2002.1166717}

\bibitem{Han2019}
Han, X., Hu, Y., Foschini, L., Chinitz, L., Jankelson, L., Ranganath, R.:
  {Adversarial Examples for Electrocardiograms}. CoRR  \textbf{abs/1905.05163}
  (2019), \url{http://arxiv.org/abs/1905.05163}

\bibitem{Mixer}
hart, M.: Just mixer. \url{https://makerhart.com}

\bibitem{Yakura2019}
Hiromu~Yakura, J.S.: Robust audio adversarial example for a physical attack.
  In: Proceedings of the Twenty-Eighth International Joint Conference on
  Artificial Intelligence (IJCAI-19). pp. 5334--5341 (August 2019).
  \doi{10.24963/ijcai.2019/741}

\bibitem{ecg-workout}
Huff, J.: {ECG} Workout-Exercises in Arrhythmia Interpretation. Wolters Kluwer
  (2017)

\bibitem{company:arterys}
Inc., A.: Arterys. \url{https://www.arterys.com/} (2019)

\bibitem{company:enlitic}
Inc., E.: Enlitic. \url{https://www.enlitic.com/} (2019)

\bibitem{Kachuee2018}
{Kachuee}, M., {Fazeli}, S., {Sarrafzadeh}, M.: {ECG} heartbeat classification:
  A deep transferable representation. In: 2018 IEEE International Conference on
  Healthcare Informatics (ICHI). pp. 443--444 (June 2018).
  \doi{10.1109/ICHI.2018.00092}

\bibitem{dnn-imaging-applications}
{Ker}, J., {Wang}, L., {Rao}, J., {Lim}, T.: Deep learning applications in
  medical image analysis. IEEE Access  \textbf{6},  9375--9389 (2018).
  \doi{10.1109/ACCESS.2017.2788044}

\bibitem{Kher2019}
Kher, R.: Signal processing techniques for removing noise from {ECG} signals. J
  Biomed Eng pp.~1--9 (2019)

\bibitem{Kurakin2016}
Kurakin, A., Goodfellow, I., Bengio, S.: Adversarial examples in the physical
  world. ICLR Workshop  (2017)

\bibitem{AAMI}
for the Advancement~of Medical~Instrumentation, A.: Testing and reporting
  performance results of cardiac rhythm and {ST}-segment measurement
  algorithms. In: ANSI/AAMI EC38 (1998)

\bibitem{Moody2001}
Moody, G., Mark, R.: The impact of the {MIT-BIH} arrhythmia database. IEEE
  engineering in medicine and biology magazine : the quarterly magazine of the
  Engineering in Medicine \& Biology Society  \textbf{20},  45--50 (June 2001).
  \doi{10.1109/51.932724}

\bibitem{Papernot2018}
Papernot, N., McDaniel, P., Sinha, A., Wellman, M.P.: Sok: Security and privacy
  in machine learning. In: 2018 IEEE European Symposium on Security and Privacy
  (EuroS P). pp. 399--414 (2018). \doi{10.1109/EuroSP.2018.00035}

\bibitem{Pranav2017}
Rajpurkar, P., Hannun, A.Y., Haghpanahi, M., Bourn, C., Ng, A.Y.:
  Cardiologist-level arrhythmia detection with convolutional neural networks.
  CoRR  \textbf{abs/1707.01836} (2017), \url{http://arxiv.org/abs/1707.01836}

\bibitem{Shah2019}
Shah, P., Kendall, F., Khozin, S., Goosen, R., Hu, J., Laramie, J., Ringel, M.,
  Schork, N.: Artificial intelligence and machine learning in clinical
  development: a translational perspective. npj Digital Medicine
  \textbf{2}(69), ~1--5 (2019). \doi{10.1038/s41746-019-0148-3}

\bibitem{Szegedy2014}
Szegedy, C., Zaremba, W., Sutskever, I., Bruna, J., Erhan, D., Goodfellow, I.,
  Fergus, R.: Intriguing properties of neural networks. In: International
  Conference on Learning Representations (2014),
  \url{http://arxiv.org/abs/1312.6199}

\bibitem{Tu2020}
Tu, J., Ren, M., Manivasagam, S., Liang, M., Yang, B., Du, R., Cheng, F.,
  Urtasun, R. (eds.): Physically Realizable Adversarial Examples for LiDAR
  Object Detection. IEEE (2020), \url{https://arxiv.org/pdf/2004.00543.pdf}

\bibitem{Turakhia2013}
Turakhia, M., Hoang, D., Zimetbaum, P., Miller, J., Froelicher, V., Kumar, U.,
  Xu, X., Yang, F., Heidenreich, P.: Diagnostic utility of a novel leadless
  arrhythmia monitoring device. The American journal of cardiology
  \textbf{112} (May 2013). \doi{10.1016/j.amjcard.2013.04.017}

\bibitem{Kan2016}
Xiao, K., Forte, D., Jin, Y., Karri, R., Bhunia, S.K., Tehranipoor, M.M.:
  Hardware trojans: Lessons learned after one decade of research. ACM
  Transactions on Design Automation of Electronic Systems  \textbf{22}(1)
  (2016). \doi{10.1145/2906147}

\end{thebibliography}

%\section*{Appendix}

% ハードウェアセットアップの写真画像

\section*{Appendix A: Concrete Attack Scenarios}

% \sugawara{Changed the section structure by putting "Appendix X:" at the heads.}

With the threat model laid out in section~\ref{sec:threat}, we give some concrete examples as to how certain adversaries may go about an attack.
Following the heartbeat types shown in Table~\ref{tbl:aami}, we present two heartbeat spoofing attack models and explain ways in which the resulting ECG readings can be used for medical fraud to benefit the adversary. 
 
\paragraph{\bf S-Target Spoofing Attack}

One of the specific conditions the S class heartbeats entails is Atrial Premature heartbeats~\cite{AAMI}. While individual Atrial Premature heartbeats are considered harmless, 3-4 consecutive occurrences entail potential signs of severe stress, side affects of medicine, excessive substance consumption, and potential heart failure~\cite{ecg-workout}. Patients with frequent occurrences of this arrhythmia are at times subject to anti-anxiety drugs, or heart medication to prevent further development of serious arrhythmia. The exact hardware attack executed in our experiments can be utilized by a malicious hospital personnel. A physician may be able to embed an adversarial hardware contraption between the input of the readings from the electrodes to feign consecutive S class heartbeats in patient ECGs. This will allow the physician to appoint excessive medication, leading to higher costs billed. Perfect performance of the hardware attack is not required to succeed, as only few of the attempts are required to succeed in order to raise suspicion of dangerous heart activity with occasional consecutive S class heartbeat occurrences.

\paragraph{\bf V-Target Spoofing Attack}

V class heartbeats point to Premature Ventricular Contraction~\cite{AAMI}. Premature Ventricular Contraction heartbeats indicate heart instability when occurring in succession, or in a pattern and is looked out for during periods following a heart surgery~\cite{ecg-workout}. Patients seen with frequent occurences of this arrhythmia are at times subject to antiarrhythmic medication. Malicious hospital personnel may conduct a similar attack to the S-Target Spoofing attack to display such arrhythmia occurring in patterns to warrant unnecessary application of antiarrhythmic medication during post-surgery care. Constant monitoring is required after heart surgeries, and devices may be assigned to patients after heart surgeries to ensure stability in heart activity, making this scenario especially vulnerable to our suggested attack method.

\section*{Appendix B: Adversarial Beat Generation Algorithm}

Algorithm~\ref{alg:gen} describes the algorithm used to generate {\em adversarial beats}. In each iteration, {\em adversarial beats} are optimized to maximize attack success rate, and a stricter amplitude is assigned once the perturbation reaches the threshold attack success rate. After $N$ unsuccessful iterations, the algorithm deems further constriction of amplitude difficult and aborts the training. For this experiment, we empirically set the threshold as $N=5$.

\begin{algorithm}[ht]
\footnotesize
\SetAlgoLined
\KwResult{Adversarial Beat with optimized amplitude $A$}
 initialize variables\;
 \quad amplitude $A=1.0$, counter $c=0$, status $s=\mbox{False}$\;
 set parameters $\Delta= 0.1$, $\rho$ as acceptable accuracy\;
 \While{c $<$ $N$}{
  $A\to A-\Delta$\;
  Train {\em adversarial beat} with equation~\ref{eq:optimization}\;
  \eIf{misclassification probability $>$ $\rho$}{
   $s=\mbox{True}$\;
   c = 0\;
   }{
   $A\to A+\Delta$\;
   $\Delta\to 0.5\times\Delta$\;
%   scale delta-height by 0.5\;
   $c \to c+1$\;
  }
 }
 \eIf{$s= \mbox{True}$}{
    Return trained {\em adversarial beat}\;
 }{
    Return Null\;
 }
 \caption{Adversarial Beat Generation}
 \label{alg:gen}
\end{algorithm}

\section*{Appendix C: Target Model Details}

% Details of the target model are discussed here.

\paragraph{\bf Architecture}

The model complies with the five-heartbeats classification in Table~\ref{tbl:aami}. The architecture is a series of consecutive residual blocks consisting of convolutional layers and max-pooling layers, followed by two fully-connected layers with 32 neurons each. Convolutional layers perform 1-D convolution, each with 32 kernels of size 5, as specified by Kachuee~et~al.\cite{Kachuee2018}. 
For the Max-pooling layer, size is 5 with a stride of 2. A classification accuracy of up to 93.4\% was achieved in Kachuee's original work, successfully determining heartbeat patterns.

\paragraph{\bf Dataset} 

To train the classifier, we adopt the PhysioNet MIT-BIH Arrhythmia Database~\cite{Goldberger2000, Moody2001}.
All the heartbeats are labeled with their respective heartbeat annotations, which are adapted to the beat annotation standards specified in AAMI EC57 as shown in Table~\ref{tbl:aami}. We first segment ECGs in the dataset into individual heartbeats. The resulting dataset contains segments of ECGs of individual heartbeats, each labeled as a class, resized to universal length, so that they can be inserted into the target classifier. We normalize the amplitude of waveforms values to 0--1.0. We subsampled all but one class to counteract disproportionate class distributions. Finally, the resulting dataset is then split with 80\% of the dataset used for training the classifier and the remaining 20\% used for testing the classifier and the effectiveness of our created {\em adversarial beats}.

\section*{Appendix D: Hardware Setup}

The ECG controller is realized with the Analog Devices AD8232 module which amplifies and filters the raw electrical signal. We use an off-the-shelf data acquisition device, National Instruments myDAQ, capable of acquiring/generating electrical to realize the ADC/DAC in the signal processing device shown in Figure~\ref{fig:hardware-setup}. A laptop PC is attached to the NI myDAQ device to inject pre-computed {\em adversarial beats} with proper timing. Signal addition was executed by using a commercial audio mixer~\cite{Mixer}. The final ADC component was implemented with an Arduino via their analog input port.

\end{document}